\documentclass{article}

\usepackage{amsfonts,amssymb,amsmath}
\usepackage{graphicx,graphics,xcolor}

\begin{document}
\title{
\textcolor{red}
{Correspondence Factor Analysis of Big Data Sets: A Case Study of 30 Million Words;
and Contrasting Analytics using Apache Solr and Correspondence Analysis in R}}
\author{Fionn Murtagh \\
{\tt fmurtagh@acm.org}}

\maketitle

\section*{Technical Report: Table of Contents}

\begin{itemize}
\item \textcolor{red}{
Part 1: Correspondence Factor Analysis of Big Data Sets: A Case Study of 
30 Million Words}
\item 1. Data
\item 1.1 General Considerations in Regard to High Dimensional and Massive Datasets
\item 1.2 Characteristics of the Data Studied
\item 1.3 Distribution of Terms
\item 2. Objective of Computational Efficiency: Using Aggregated Subsets to
Construct the Factor Space
\item 2.1 Introduction
\item 2.2 Projection into the Factor Space
\item 2.3 The Practical Benefit of a Selected Word Set
\item 2.4 Conclusion
\item \textcolor{red}{Part 2: Data Analysis of Recipes: 
Contrasting Analytics using Apache Solr and Correspondence Analysis in R}
\item 3. About the Data; followed by Correspondence Analysis of All Recipes 
using a 247 Word Set
\item 4. Search Using Solr
\item 4.1 Setting Up Data File and Its Description
\item 4.2 Running the Server and Updating the Index
\item 4.3 Querying and Other Operations
\item 4.3.1 Web Browser User Interface
\item 4.3.2 Command Line Querying
\item 5. Consistency of Solr Searches
\item References
\item Appendix 1: Sample Recipe
\item Appendix 2: 247 Ingredients Used as Attributes in the 
Correspondence Analysis
\item Appendix 3: Alternative Presentation of Plot
\item Appendix 4: Correspondence Analysis for Singular/Plural Association and           
Potentially for Disambiguation
\end{itemize}

\begin{abstract}
We consider a large number of text data sets.  These are cooking recipes.
Term distribution and other distributional properties of the data are 
investigated.  Our aim is to look at various analytical approaches which 
allow for mining of information on both high and low detail scales.  Metric 
space embedding is fundamental to our interest in the semantic properties of
this data.  We consider the projection of all data into analyses of aggregated 
versions of the data.  We contrast that with projection of aggregated versions 
of the data into analyses of all the data.  Analogously for the term set, 
we look at analysis of selected terms.  We also look at inherent term 
associations such as between singular and plural.  In addition to our use
of Correspondence Analysis in R, for latent semantic space mapping, we also 
use Apache Solr. Setting up the Solr server and carrying out querying is 
described.  A further novelty is that querying is supported in Solr based 
on the principal factor plane mapping of all the data.  This uses a bounding 
box query, based on factor projections.  

\end{abstract}

\section*{Part 1: Correspondence Factor Analysis of Big Data Sets: A Case Study of 
30 Million Words} 

\section{Data}

\subsection{General Considerations in Regard to High Dimensional and Massive Datasets}

Very high dimensional (or equivalently, very low
sample size or low $n$) data sets, by
virtue of high relative dimensionality alone, have points mostly lying
at the vertices of a regular simplex or polygon \cite{hall,ref08}. 

In \cite{morin}, the following issues are posed.  Firstly, what should we do, 
for our analytics, when we have thousands of documents?  Correspondence Analysis 
outputs (and this is the same for other kindred methods, like LSA, latent semantic 
analysis; PLSA, probabilistic adaptation of LSA using a mixture of
multinomial distributions;  
Factorial Correspondence Analysis (CA); Kohonen maps; or clustering 
methods) are very huge (much more than the original dataset).  What we must therefore
do is to display the output in an ``intelligent'' and friendly way.

\subsection{Characteristics of the Data Studied}
\label{sectchar}

We explore the following.  

\begin{itemize}
\item We have 306 collections, labelled by ``Recipe via Meal-Master (tm) v8.05'',
each with, in principle, 500 recipes.
\item (An additional set of 20 collections, relating to ``v8.06'', were considered 
as different, and set aside from this work.)
\item There is one exception here: one of the collection files
had two recipes less than the usual 500 complement.
\item Hence we are dealing with 306 $\times$ 500 recipes (less 2), 
i.e.\ 152,998.   
\item Appendix 1 shows a sample recipe.  
\item This data, in text, contained: 5,948,739 lines of text; 30,199,625 
(whitespace-demarcated) words; and 206,993,672 characters.  
\item On the 306 recipe collections, comprising 152,998 recipes, we used
our term extractor.  This listed all terms of one character or more, having 
first removed punctuation and numeric character.  We will use the term ``words''
for what was extracted.  All upper case had been first set to lower case. 
Abbreviations and acronyms showed up as words, some 
character combinations 
did so also following punctuation removal, and all forms of stemming were 
retained.

Various misspellings are in the given data.  In collection mm066001.mmf, 
there is: ``Grate the zuccini into a calendar (sp)...''.  What is intended
is ``zucchini'' and ``colander''.  Interestingly there are 57 times that 
``zuccini'' appears, and once that ``zuccinis'' appears.  

While spelt correctly, ``zucchini'' is present 7155 times, and ``zucchinis'' is 
present 56 times.  There are various other incorrect words present, with 
one occurrence each: ``azucchini'', ``brushingzucchini'', ``lambzucchini'',
``ofzucchini'', ``pzucchini'', ``zucchinia'', ``zucchiniabout'', 
``zucchinii'', ``zzzucchini''.

Since we take the data as such, notwithstanding such inaccuracies, it is 
to be noted that the correct spellings are predominant, and that the 
misspellings only show up if quite atypical features in the data (such 
as rare occurrence) are looked into.  Let us formulate this as a 
working principle, or perhaps even a working hypothesis that we have 
verified in all cases that we have looked into.
In the case of large, or very large, numbers of 
occurrences of entities, the syntactically correct form will predominate.  
  
\item From the 152,998 recipes we obtained 101,060 unique words. 
\item In this list, ranked by decreasing frequency of occurrence, the 
5000th ranked word had 162 occurrences; the 15,000th ranked word had
18 occurrences; and at the 56,229th ranked word (56\% through the total list
of 101,060 ranked words), the frequency of occurrence
became thereafter (for higher ranked words), 1.  
\item The top frequencies of occurrence are as follows:
\begin{verbatim}
and 890294
the 842542
to 531633
in 469003
with 313588
of 281320
recipe 280035
or 247912
ts 238309
for 226514
until 224481
add 207146
tb 200221
from 193659
minutes 190880
by 190815
on 183287
salt 162811
yield 162082
servings 161098
meal 155040
master 152980
via 149162
title 148802
tm 148766
categories 148612
into 145565
sugar 138670
heat 131846
water 130211
pepper 127901
oil 122997
chopped 119530
over 115212
butter 112329
at 106300
sauce 102881
is 101060
com 99841
flour 96793
stir 91580
mixture 91484
about 91279
it 89491
cook 87097
pan 86314
oz 83908
garlic 82776
cream 81633
place 80825
cheese 78986
bowl 76209
mix 76182
chicken 75807
onion 71516
cut 70808
lb 68376
posted 67769
fresh 66125
baking 64731
\end{verbatim}
\item The very final ranked words were as follows:
\begin{verbatim}
zwiebelfleisch 1
zwirek 1
zws 1
zygielbaum 1
zyla 1
zylman 1
zza 1
zzwc 1
zzzingers 1
zzzucchini 1
\end{verbatim}

The second last here arose from this entry in a recipe:
``Title: Wing Zzzingers''.    The last word 
comes from this entry in a recipe:  ``Add zzzucchini,
peppers and onion.''.   
\end{itemize}

\subsection{Distribution of Terms}

A power law (see \cite{mitz})
is a distribution (e.g.\ of frequency of occurrence)
of the form $x^{-\alpha}$
where constant, for the data,  $\alpha > 0$; and an exponential law is
of the form $e^{-x}$.  For a power law, $P(x > x_0) \sim c x^{-\alpha}$,
$c, \alpha > 0$.  A power law has heavier tails than an exponential
distribution.  In practice $0 \leq \alpha \leq 3$.  For such values,
$x$ has infinite (i.e.\ arbitrarily large)
variance; and if $\alpha \leq 1$ then the mean of $x$ is
infinite.

The density function of a power law is $f(x) = \alpha c                         
x^{-\alpha - 1}$, and so $\ln f(x) = - \alpha \ln x  + C$, where $C$ is a
constant offset.  Hence a log-log plot shows a power law as linear.  Power
laws have been of great importance for modelling networks and other
complex data sets (see \cite{eiron,newman}).

\begin{figure}  
\centering
\includegraphics[width=14cm]{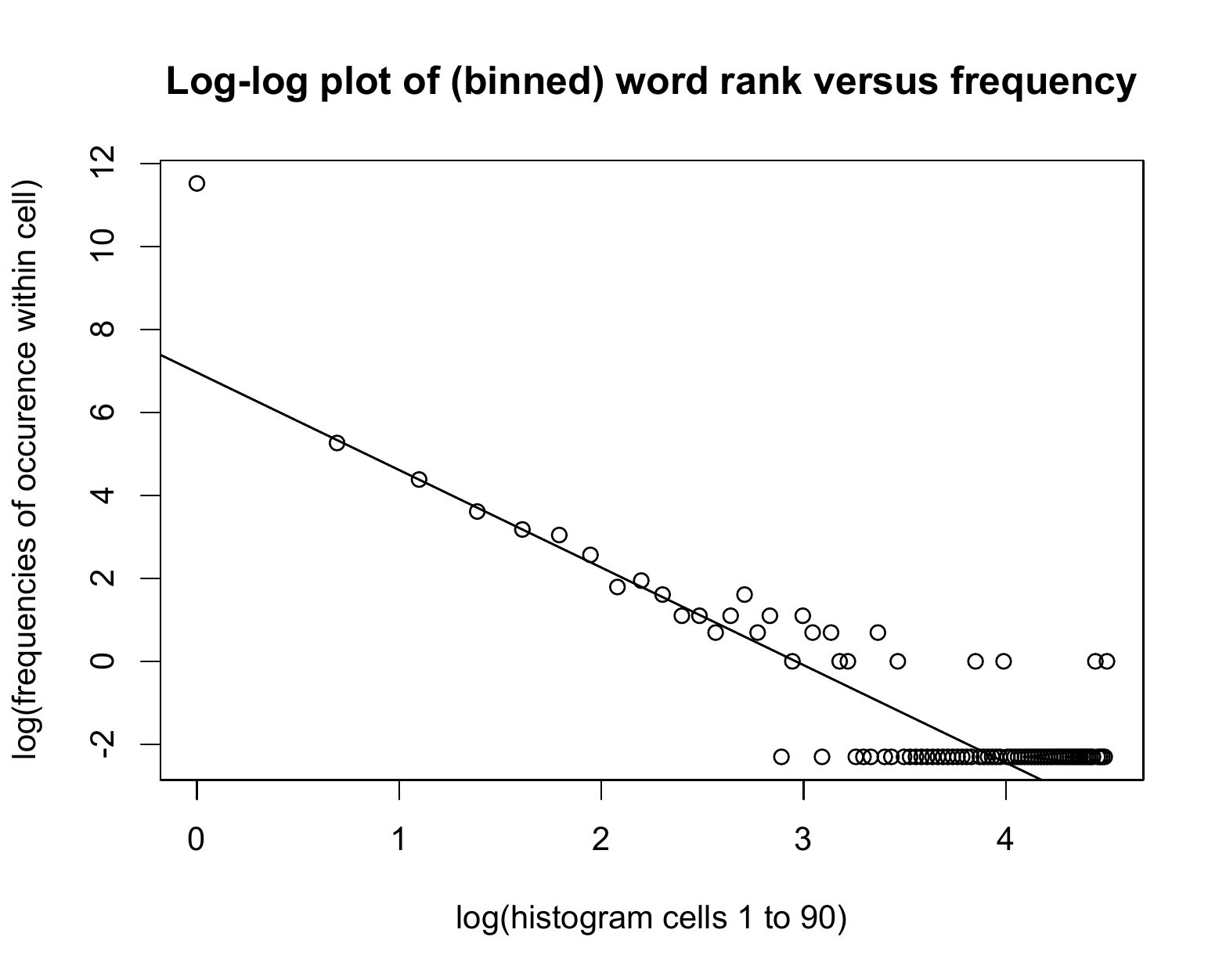}
\caption{Log-log plot of histogram of word ranks versus frequency of occurrence.
This uses 101,060 unique words of one character or more.  Summarized in this 
plot there are 30,199,625
words in total, culled from 152,998 recipes.  The line of best fit in the 
linear regime is used to calculate the power law expression for this data.}
\label{fig1}
\end{figure}

Figure \ref{fig1} shows a plot of rank (most frequent, through to least frequent, 
the latter being a very large number of words that occur once only) 
against the value of the frequency of occurrence.  To understand the plot, 
we use log-log scaling.    

In a very similar way to the power law properties
of large networks (or file sizes, etc.) we find an approximately linear
regime, ending (at the lower right) in a large fan-out region.
The slope of the linear region characterizes the power law.  We find that
the probability of having more than $x$ occurrences per word to be approximately
$c/x^{-2.3515}$ for large $x$ (since we find the slope of the fitted line 
in Figure \ref{fig1} to be $-2.3515$). 


\section{Objective of Computational Efficiency: Using Aggregated Subsets to 
Construct the Factor Space}

\subsection{Introduction}

Just how well can Correspondence Analysis carry out the Euclidean, factor 
determining mapping,
if recipes are first aggregated together?  One reason for this question is
computational time.  
For analysis of an $n \times m$ data set, with $m$ words and $n$ recipes or
recipes collections, then the dual spaces means that we analyze either the 
data set or its transpose.  Eigen-reduction is a cubic process.  The computational
time for analysis of the $n \times m$ data set is $O(m^3)$, or for its transpose,
$O(n^3)$.  For a selected word set, if we can carry out the main analysis 
on a smaller $n$, so much the better for us.  

Aggregation is just 
concatenating the recipes.  In the entire set of recipes, the number of words 
(to repeat: of one
or more characters in length; punctuation deleted; upper case set to lower case;
numerical and accented characters ignored) was 101,060.  For initial assessment,
we took the top-ranking 1000 words.  So this illustrative case, using just
500 recipes, has the recipe set crossed by this 1000-word set, with the 
frequencies of occurrence tabulated.   

We aggregated the 500 recipes into 5 recipe-sets.  These comprised recipes 1--100,
101--200, 201--300, 301--400, and 401-500.   Each of these 5 recipe-sets had the
frequencies of occurrences on the 1000-word set.   

Here we analyze two cross-tabulation tables, of dimensions respectively $500 \times 
1000$, and $5 \times 1000$.  By construction of the latter, the column sums are 
identical.  Then due to the use of profiles, the average profile is the same in each 
case.   If we represent the data matrix in frequency terms (i.e.\ the frequencies 
of occurrences each divided by the grand total), the matrix can be denoted 
$f_{IJ} = \{ f_{ij} | i \in I, j \in J \}$ where $I$ is set of recipes, or of 
recipe-sets, $J$ is the word-set, $i$ is a recipe or a recipe-set, and $j$ 
is a word)  the column marginal distribution is $f_J = \{ f_j | j \in J \}$,
or in summation terms, $\sum_{i \in I} f_{ij}$.  

In Correspondence Analysis, 
the centre or origin of the Euclidean factor space is given by $f_J$ in 
the recipe space, and by $f_I$ in the word space, such that we have the 
following view: starting with a probability distribution $f_{IJ}$ we want to 
explore how it differs from the product of marginal distributions, $f_I f_J$, 
and this in the $\chi^2$ metric of centre, the product $f_I f_J$.  This is 
a decomposition of inertia of clouds in recipe and/or in word space. 
We have: $M^2(N_I(J)) = M^2(N_J(I) = \| f_{IJ} - f_I f_J \|^2_{f_I f_J} 
= \sum_{i \in I, j \in J} (f_{ij} - f_i f_j)^2 / f_i f_j =
\sum_{i \in I} f_i \| f^i_J - f_J \|^2_{f_J} =
\sum_{j \in J} f_j \| f^j_I - f_I \|^2_{f_I}$ where $f^i_J = \{ f_{ij} / f_i \}$.  

\begin{figure}  
\centering
\includegraphics[width=14cm]{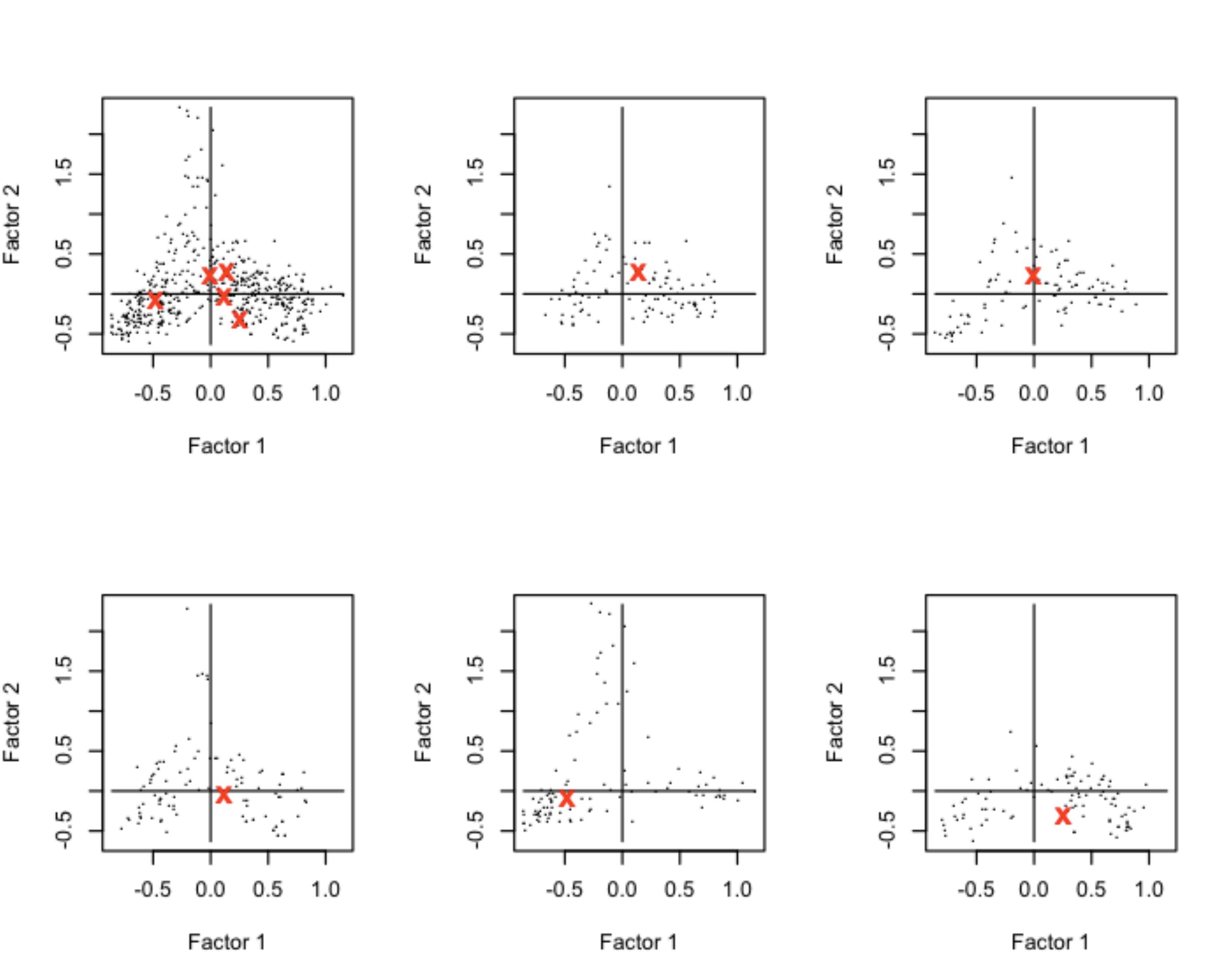}
\caption{Upper left: projection on principal factor plane of 500 recipes, located 
at the dots; and the 5 recipe-sets, located at the centres of the x symbols.  
The analyses were carried out separately.  Just to see how the recipe-sets 
are at the centres of gravity of their associated recipes, displayed here in
planar projections, the successive panels show just the recipe-set projected, 
and the associated 100 recipes, projected as dots.}
\label{cawithagg}
\end{figure}

Figure \ref{cawithagg} shows the principal factor display resulting from the 
Correspondence Analysis of (1) the $500 \times 1000$ recipes times words 
data, and (2) the $5 \times 1000$ recipe-sets times words data.  Note again 
that the analyses (the eigen-reductions -- determining the factors) were 
completely independent.  

The results of Figure \ref{cawithagg} can be explained in the following 
terms.  We have a set of 500 items (the recipes) with quantitative 
description in a 1000-dimensional space.  We have a derived, through 
concatenating, sets of 5 items in the same 1000-dimensional space.  So 
the resultant Euclidean mappings illustrate well the shared provenance 
in the two cases.  The word analysis is less clear-cut.  Given the 
inherent dimensionality of the Euclidean, factor space is min($n-1, 
m-1$), we have, on the one hand 1000 words projected into a 499-dimensional
space, and on the other hand 1000 words projected into a 4-dimensional
space.   The locations of a given word in these two spaces are not 
directly related.

\subsection{Projection into the Factor Space}
\label{sectproj}

We investigate supplementary elements to expedite the computation.  Carry out 
analysis in a small dimensional space, that nonetheless takes account of 
all relationships in an aggregated way.  Then complete the analysis on all
the data by projecting into the Euclidean, factor space.  

We took the 306 collected sets of, each, 500 recipes (save for the case of 2 
recipes short of this in one such collected set: hence we were dealing with 
152,998 recipes).   The number of words found in this data (to repeat: of one 
or more characters in length; punctuation deleted; upper case set to lower case; 
numerical and accented characters ignored) was 101,060.  For initial assessment,
we took the top-ranking 1000 words.   


The following principles and practices of Correspondence Analysis will now be availed of
(see \cite{leroux1,leroux2,ref08888}):  

\begin{itemize}

\item In Correspondence Analysis, all frequency of occurrence values are divided 
by the grand total (so that allows the viewpoint of empirical probabilities). 
If the row $i$ mass is then $w_i$, the row $i$ profile is the row vector $x_i$
with each element divided by $w_i$.  Now consider the 360-set, which is just 
all 500 recipes, $x_i$, summed.   The profile of this set is $\sum x_i /
\sum w_i$ where $i = 1, 2, \dots , 50$.  If one recipe is a multiplicative
constant of another (i.e.\ elementwise, for each word), then their profiles 
are the same. 

\item The dual space relations allows supplementary rows or columns to be projected, 
after the fact, into the analysis. 

(In the Latent Semantic Indexing context, ``folding in'' of newly presented
rows or columns, that can be considered as a row or column set that is 
juxtaposed to the original $I \times J$ matrix, is described in 
\cite{deerwester}).

\item Profiles are analyzed in Correspondence Analysis, meaning that row vector, 
and column vector, values have been divided by the associated row, or column, total. 
Consider the row or column total as a mass, as is done in this context.   By having 
all values first divided by the grand total of the rows/columns array, we have all 
values bounded by 0 and 1.  (Note that we require non-negative values for the mass
to be workable for us here.)  In this way our 306-set of (500) recipes is the 
{\em weighted mean} of the associated recipes

To see this, consider the given frequency of occurrence data divided by the grand total 
(over all recipes and words), on the set of recipes, that we will represent by matrix $x$.  
Consider the 500 recipes that are associated with one of the 306-sets.  Without loss of 
generality, call them $x_{1j}, x_{2j}, \dots , x_{500,j}$.  We will denote one such recipe by 
$x_{kj}$ and just right now we are most interested in the set $\{ x_{kj} | k = 1, 2, \dots , 500;
j \in J \}$. $J$ is the set of words.  Then the {\em profile} of recipe $k$ is $x_{kj} / 
\sum_j x_{kj}$.    

Now consider the 360-set.  It is the aggregation (concatenation) of the recipes. 
So, it is $\sum_{k = 1}^{50} x_{kj}$.   Its profile is $ \sum_{k = 1}^{50} x_{kj}  /
\sum_{k,j} x_{kj}$.  

\end{itemize}

\begin{figure}  
\centering
\includegraphics[width=14cm]{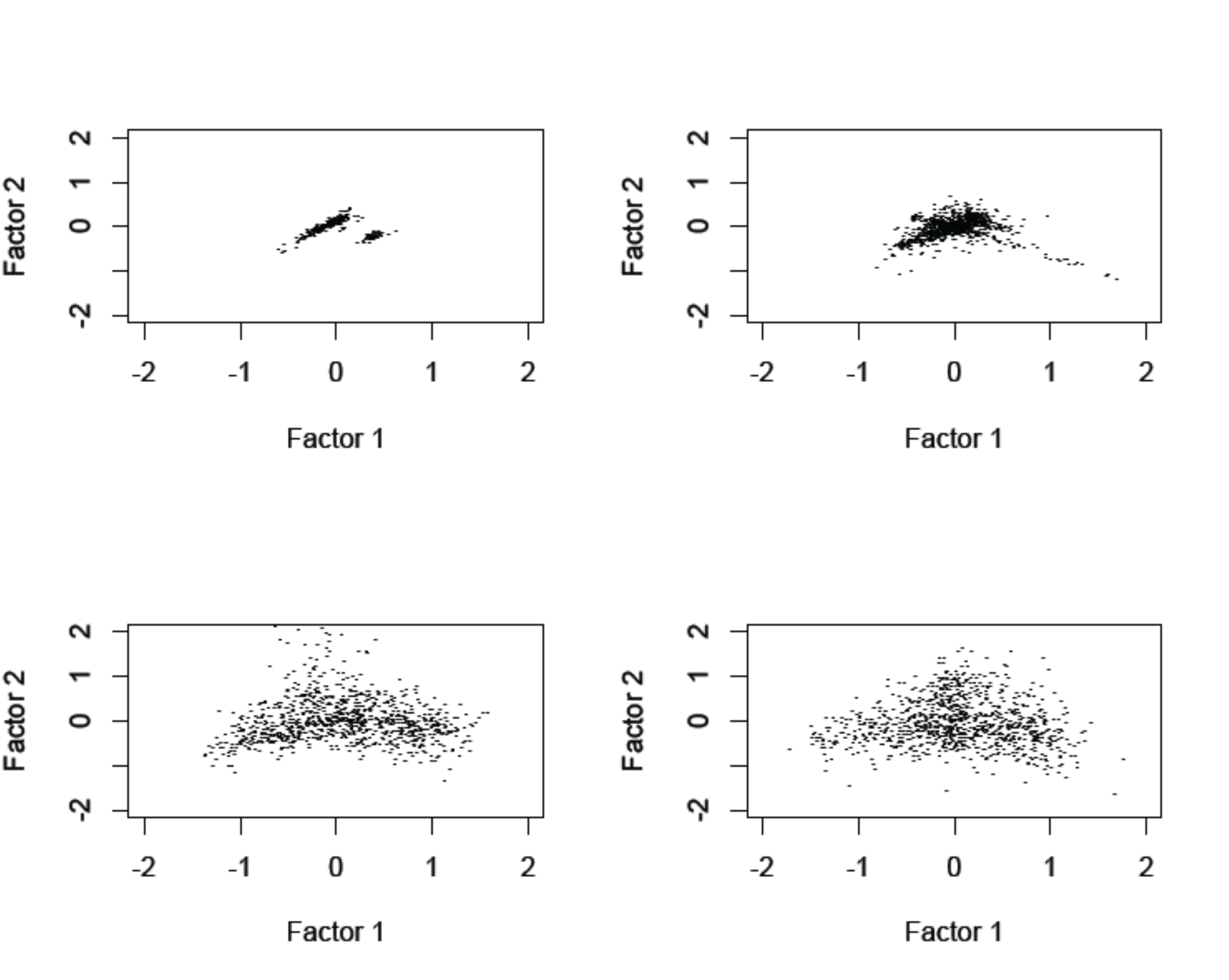}
\caption{Upper left: 306 collections of 500 (in one case, 498) recipes.  
Upper right: the 
top ranked 1000 words 
used for the 306 collections of recipes.  The upper right is the dual space 
of the upper left.  
Lower left: the 1000 words for one set of 500 recipes.   Lower right: the 
1000 words for another
set of 500 recipes.}
\label{proj1fig}
\end{figure}

Figure \ref{proj1fig} presents an initial look at the use of the top ranked 
words (in terms of 
frequency of occurrence in the entire data collection, hence: 306 collections of recipes, 
or 152,998 recipes). 

\begin{figure}  
\centering
\includegraphics[width=14cm]{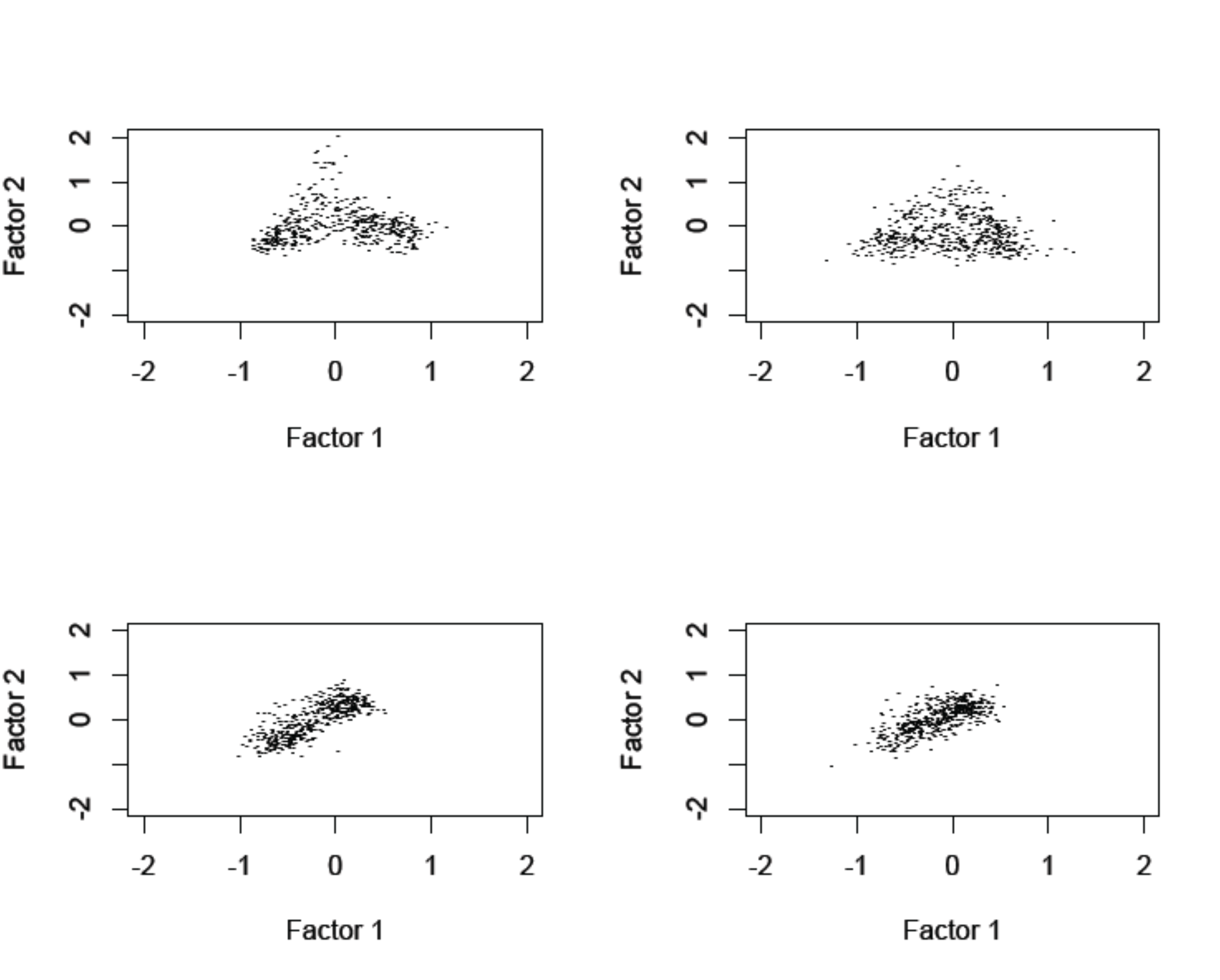}
\caption{Upper left and right, respectively: the two sets of 500 recipes as used in Figure
\ref{proj1fig}.  Lower left and right: these two sets of 500 recipes are projected into the
analysis of the 306 recipe collections.  What enables this to be accomplished is that
the same 1000-word set is used in all cases here.}
\label{proj2fig}
\end{figure}

In Figure \ref{proj2fig} we look at the potential for handling massive data sets by 
projecting into a ``primary'' analysis.   We have this analysis here on the 
$306 \times 1000$ recipe collections times 1000 highest ranked words.   We project
into the analysis the first $500 \times 1000$ recipe set, and the second $500 \times 
1000$ recipe set.   

The scheme used is as follows.   We take as a main matrix the one that crosses 306 rows 
with 1000 columns.  Then 
we have a supplementary row set, of 500 rows crossed with 1000 columns.  Finally we have 
a further supplementary row set, of 500 rows crossed with 1000 columns.  The supplementary 
rows are projected into the Correspondence Analysis of the $306 \times 1000$ matrix, post factum.  

We note that it is best to carry out the eigen-reduction on the smaller of the row set
and the column set.  Hence, relative to how we have presented our processing steps above,
in fact we worked on the transposed matrices.  (For the $306 \times 1000$ matrix, diagonalization
is carried out on a $306 \times 306$ matrix.) 

\begin{figure}  
\centering
\includegraphics[width=14cm]{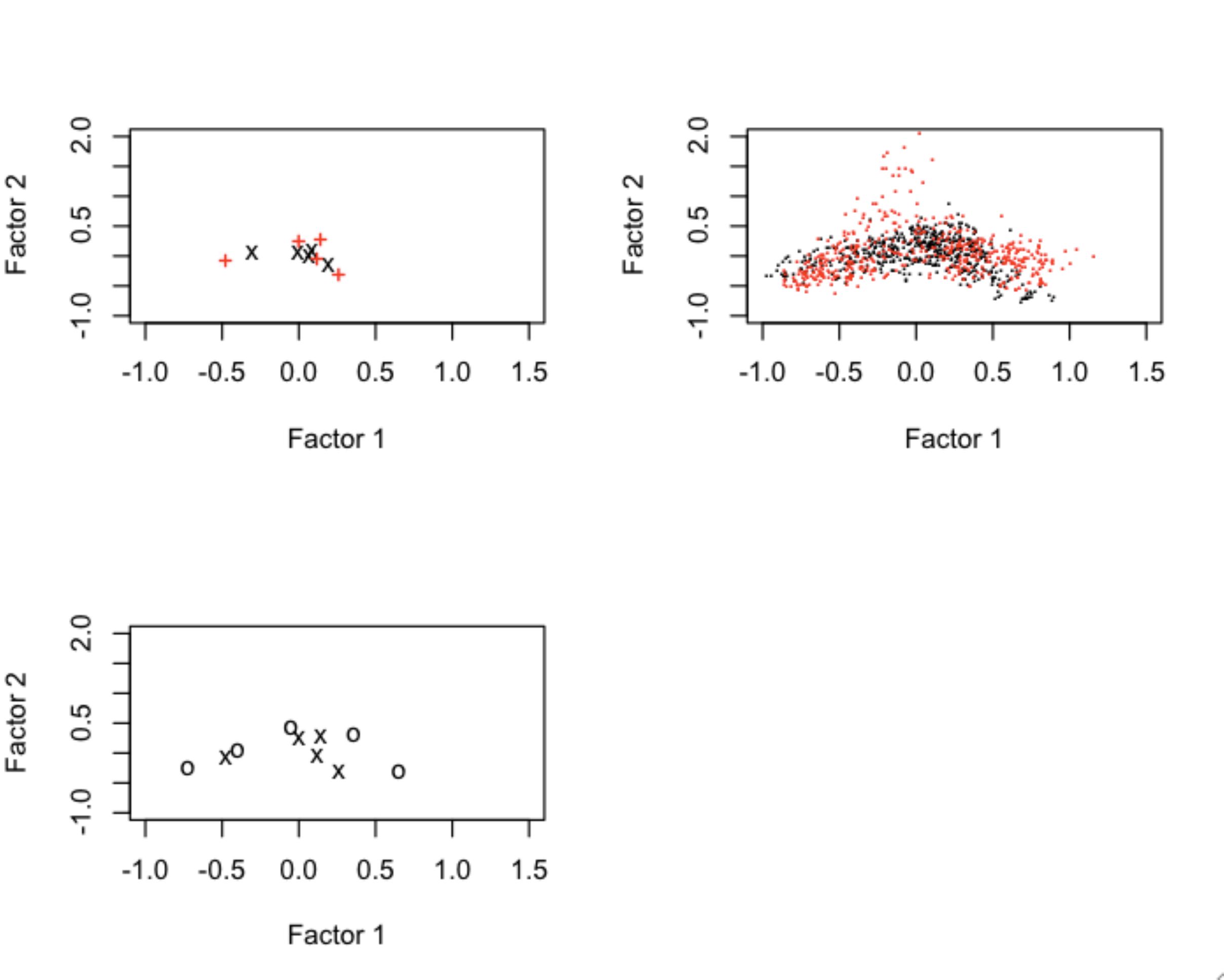}
\caption{Upper left: red ``+'' are principal factor plane projections of 
the 5 recipe-sets, in the $5 \times 1000$ dataset, and black ``x'' are the 
projections, relative to principal factor plane projection of using the 
original $500 \times 1000$ data.  Upper right: a black dot, ``.'', represents
projections as supplementary elements of the 500 recipes, relative to the
derived 5 recipe-sets; and a red dot, ``.'', represents the active 
Correspondence Analysis of these $500 \times 1000$ data.  So the black dots
are efficiently determined, but the red dots show the ``ground truth''.  
Lower panel: different set of 5 recipe-collections, as used previously (e.g.\
upper left panel here), and then independently defined in order to enhance
having ``like with like'' in the same recipe-collection.  
Note that in all cases we are using a 2-dimensional (albeit optimal) display.}
\label{bothways-diff5}
\end{figure}

In Figure \ref{bothways-diff5}, in the upper left we show with ``x'' (i) the 
projection as supplementary elements of a $5 \times 1000$ set of data derived
from a $500 \times 1000$ set of data; and shown with a red ``+'' (ii) the 
projections of the active Correspondence Analysis of the $5 \times 1000$ set of data.   

The $500 \times 1000$ set of data 
was the set of 500 recipes used before -- in fact the sequentially first of 
these batches of recipes (and as indicated there were 306 of these batches).  
The word set is 1000, being the top-ranked words in the entire word-set
derived from all recipes.   From the $500 \times 1000$ set of data, we
took the first 100 recipes, then recipes 201--300, 301--400, and 401--500.
We aggregated the recipes, which is equivalent to concatenating the recipes. 
This provided a frequency of occurrence cross-tabulation of $5 \times 1000$. 

In Figure \ref{bothways-diff5}, upper right, we show with a dot, ``.'', 
(i) the 500 recipes with their projections found from using the $500 \times 1000$ 
data as supplementary elements relative to the $5 \times 1000$ data.  We show 
also, using red dots, ``.'',  (ii) the active Correspondence Analysis of the $500 \times 1000$ 
data.  

In Figure \ref{bothways-diff5}, lower panel, there is, with an ``x'', (i)
the projections of the active Correspondence Analysis of the $5 \times 1000$ set of 
data, as in the upper left.  Then, with a ``o'', there is (ii) the same analysis 
on a matrix of the same dimensions, but this time with the 5 concatenated sets of 
recipes being based not on the given sequential order, but rather on the first
factor projections of the full $500 \times 1000$ matrix.   This was done in order
to assess a different derived frequency of occurrence set of data.

Arising out of this discussion, we will draw a balance sheet on 
projections, through assessment of similarity of outcomes.  
We use the sums of squared Euclidean distances, taken from the principal plane
factor projections, which are Euclidean.  

\medskip

\noindent
{\bf Quality of results based on sum of squared Euclidean distances.}

\begin{enumerate}

\item
{\bf Concatenated recipes projected onto full recipe set, versus concatenated recipes:}
$5 \times 1000$ projected onto $500 \times 1000$, and  active analysis of $5 \times 1000$:   
{\bf 0.03306077}

\item
{\bf Full recipe set projected onto concatenated recipes, versus full recipe set:}
$500 \times 1000$ projected onto $5 \times 1000$,  and  active analysis of $500 \times 1000$:  
{\bf 0.283562}
\end{enumerate}

Next, $5 \times 1000$ is ``ordered'' i.e.\ based on a chosen sequencing and not an arbitrarily given one.  The chosen sequencing is that of projections on the first factor.  So,
in the foregoing experiment, the 5 groups of recipes, each of 100 recipes, had their
sets of 100 recipes in the given, arbitrary, sequence.  Now, the 5 groups of recipes,
each of 100 recipes, have their sets of 100 recipes in a somewhat clustered sequence that
is provided by these recipes' projections on the first factor.  

\begin{enumerate}
\item
{\bf Concatenated clustered recipes projected onto full recipe set, versus concatenated clustered recipes:}
$5 \times 1000$ projected onto $500 \times 1000$,  and  active analysis of $5 \times 1000$:  
{\bf 0.9963667}

\item
{\bf Full recipe set projected onto concatenated clustered recipes, versus full recipe set:}
$500 \times 1000$ projected onto $5 \times 1000$,  and  active analysis of $500 \times 1000$:  
{\bf 1.097960}
\end{enumerate}

\medskip

What is noteworthy here is that the ``ordered'', i.e.\ somewhat more clustered 100-strong 
recipe-sets, preform less well 
than arbitrarily-given 100-strong recipe sets.   (Cf.\ latter two results
above vis-\`a-vis the former two.)  We draw the conclusion
that the lesser coherence associated with the latter is better when it comes to 
projecting other recipes into this space.   

We see a certain parallel between this 
outcome and the importance of incoherence in compressed sampling in signal processing. 
Such incoherence is when \cite{starck} (p.\ 278), there should be as much spread as 
possible between the sensing vector, on the one hand, and on the other hand, 
the ``sparsity atoms'' or basic components that we can use for the signal to be acquired.

We also have an indication from the above that the bigger the better the analysis, 
in terms of
observations, when it is a matter of projecting into the factor space.  (Cf.\ first 
result above, out of the four, being better than the second result.) 

We conclude: the bigger the set of observations that we can work on, so much the better.
Also 
if we do need to work on a limited, representative set of aggregates, then lack of 
coherence (i.e.\ lack of clustering) is best. 

\subsection{The Practical Benefit of a Selected Word Set}
\label{sectwordset}

We use again the first 500-recipe set. 
With the word set, as before, derived from the entire set of recipes, we look at (i) the
top ranked (i.e.\ most frequently occurring) 1000 
words, (ii) the 42,052 word list such that word lengths are 
greater than 2, and (iii) the full word set, with 101,060 words.  
In each case, we therefore used the 500 recipes cross-tabulated with the
words in terms of frequency of occurrences.  Recall again that the 
word list used was derived from the entire set of recipes (and not just 
these 500).   

\begin{figure}  
\centering
\includegraphics[width=14cm]{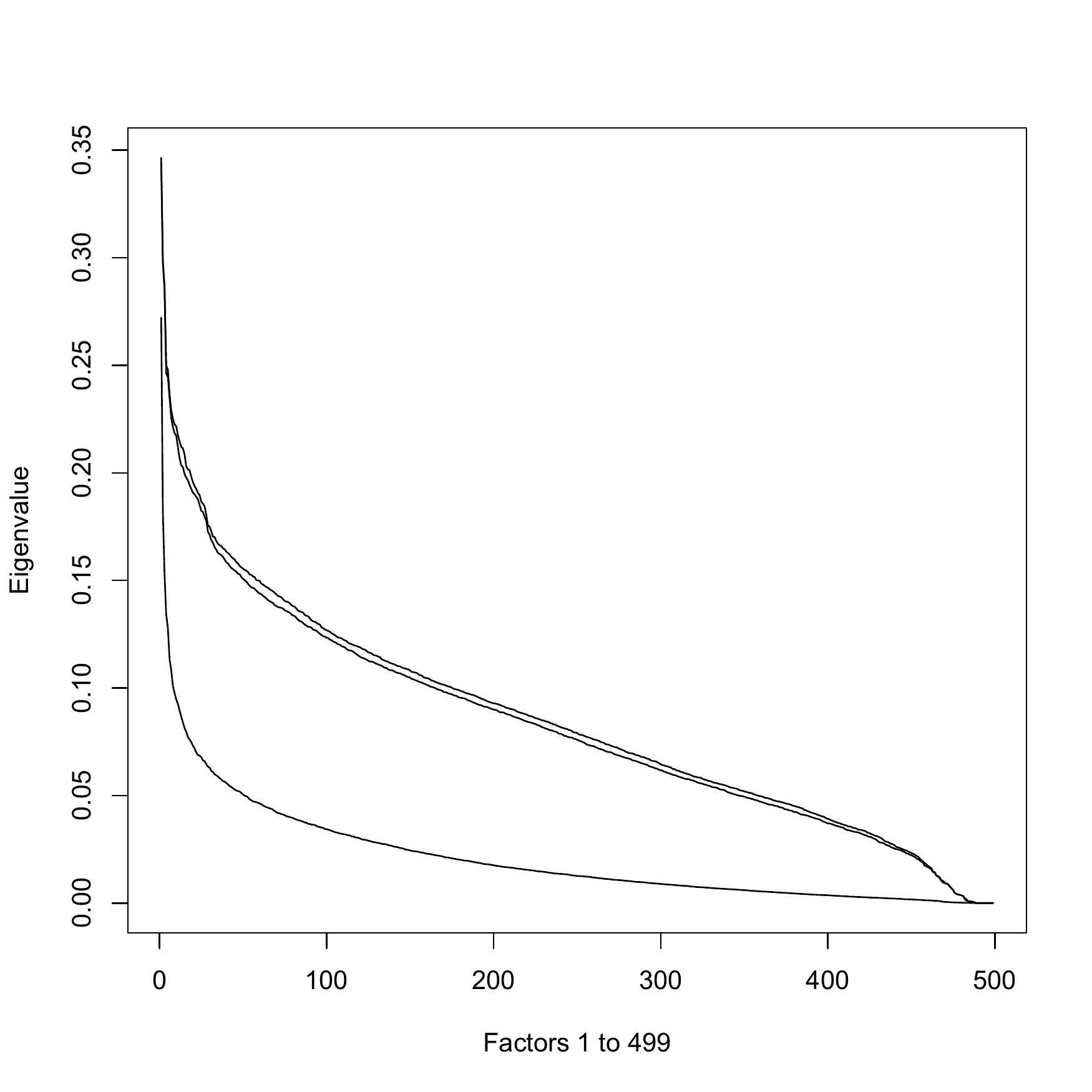}
\caption{Eigenvalues from Correspondence Analyses of a set of 500 
recipes (the very first of the
306 recipe-sets).  Top curve -- more or less fully superimposed: using
42,052 words and using the entire word set of 101,060 words.  Bottom 
curve: using 1000 words.}
\label{figevals1}
\end{figure}

\begin{figure}  
\centering
\includegraphics[width=14cm]{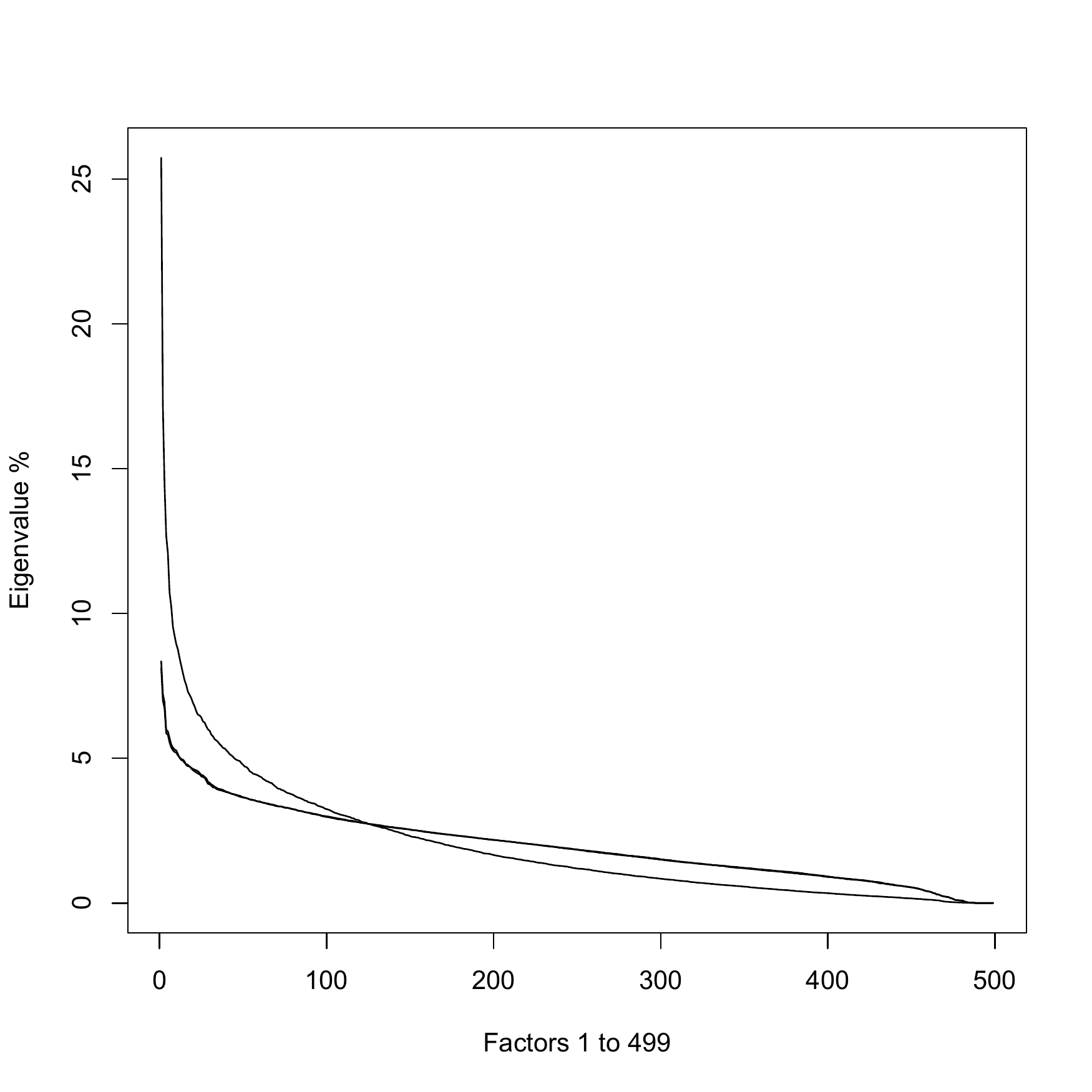}
\caption{Eigenvalues in percentage terms.  Cf.\ Figure \ref{figevals1}
displaying the eigenvalues themselves.   Here the top curve relates to 
the set of 1000 words, and the bottom -- 
more or less fully superimposed -- curves relate to the 
42,052 word set and using the entire word set of 101,060 words.}
\label{figevals2}
\end{figure}

Figures \ref{figevals1} and \ref{figevals2} portray the rapid falloff in 
inherent embedding dimensionality, based on the word set, and associated 
attribute dimensionality.   

In Figure \ref{figevals1}, the (effectively superimposed) 42,052 and 
101,060 word sets are on top, and the 1000 word set on the bottom.  
The reason for the two top curves being superimposed is the following.
The full word set is just the words that occur twice or once in the 
recipes.  Hence they are very rare words.  Hence, too, while the 
500 recipes here are characterized by, on the one hand, 42,052 words, 
and on the other hand, 101,060 words, nonetheless the extra data is
extremely sparse -- consisting mostly of zero values.  It is small 
wonder therefore that these two datasets, relating to words with 
more than two occurrences, and all words, give practically the same
outcome.  

The top 8 actual eigenvalues are as follows, if we choose, on the basis
of Figure \ref{figevals1} to select, e.g.\ 8 factors as bearing the  most 
important information: 

\begin{verbatim}
xx1c$evals[1:8]
[1] 0.3441775 0.2973126 0.2854613 0.2463559 0.2445217 0.2348758 
    0.2255849 0.2212731
x1c$evals[1:8]
[1] 0.2722012 0.1816320 0.1526613 0.1341513 0.1278123 0.1133759 
    0.1084320 0.1009735
\end{verbatim}

Our preference is to focus our interest on just the first three factors, given
these eigenvalues.


\subsection{Conclusions}

In section \ref{sectchar}, we observed that: 
In the case of large, or very large, numbers of 
occurrences of entities, the syntactically correct form will predominate.  

In section \ref{sectproj}, we concluded that: 
The bigger the set of observations that we can work on, so much the better.
Also 
if we do need to work on a limited, representative set of aggregates, then lack of 
coherence (i.e.\ lack of clustering) is best. 

Finally, in section \ref{sectwordset}, we concluded that:
Large, sparse word sets lead to a similar outcome; and a well selected, smaller
sized word set is important for best, i.e.\ smallest, reduced dimensionality 
mapping.  

\section*{Part 2: Data Analysis of Recipes: 
Contrasting Analytics using Apache Solr and Correspondence Analysis in R}

\section{About the Data; followed by 
Correspondence Analysis of All Recipes using a 247 Word Set}

The 152,998 recipes were found to have 
101,060 unique words of 1 character
or longer, having set all upper case to lower case, and having ignored 
punctuation, numeric and accented characters.   The distribution of terms
follows a power law (Zipf's law): the probability of having more than $x$ 
occurrences per word was found to be approximately $c/x^{-2.3515}$.  

In total we have 152,998 recipes.  
A set of 247 recipe ingredients was assembled.  They were chosen from a 
word list (unique words) drawn from the recipes and ordered by decreasing 
frequency of occurrence.  There were just 122 recipes that did not contain
at least one of these 247 ingredient terms.  (That could be due to use of
less common ingredients, not figuring in our list of 247; or misspellings, 
of which there were a considerable number; or some unusual text instead of a
recipe.  An example of the latter was a short overview of appropriate wines
for types of food.)

\begin{figure}
\begin{center}
\includegraphics[width=16cm]{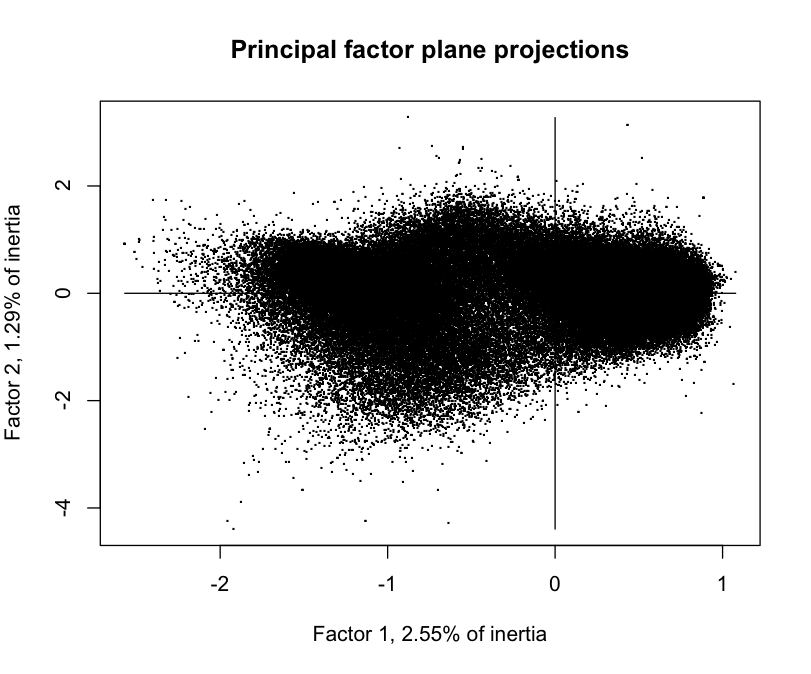}
\end{center}
\caption{Principal factor plane showing projections of 152,998 recipes,
each represented by a dot.}
\label{F1F2rec}
\end{figure}

Figure \ref{F1F2rec} shows the projections of recipes on the principal 
factor plane.  

\begin{figure}
\begin{center}
\includegraphics[width=16cm]{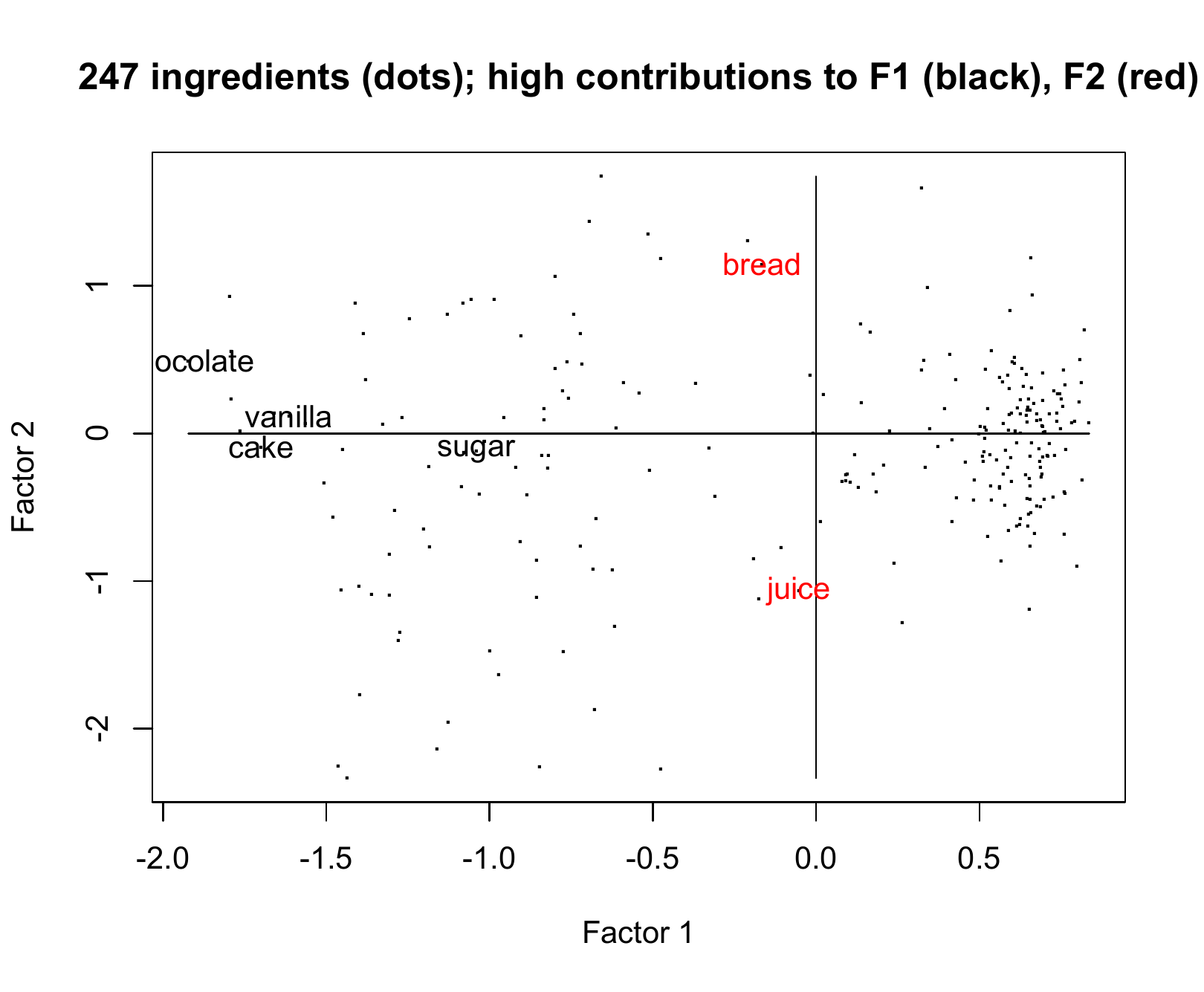}
\end{center}
\caption{Principal factor plane showing projections of 247 ingredients,
using a dot to represent projection.  Strongest contributions to factor
1: ``sugar'', ``chocolate'', and a little less, ``vanilla'', ``cake''.  
Strongest contributions to factor 2: ``juice'', ``bread''.} 
\label{F1F2ing}
\end{figure}

Figure \ref{F1F2ing} shows the ingredients used, projected onto the
principal factor plane.  The ingredients with the strongest contributions
to these factors are noted.  Looking further, we found that the strongest 
contribution by ingredients to factor 3 is ``chocolate''; and the strongest
contribution to factor 4 is ``cheese''.

\section{Search using Solr}

\subsection{Setting Up Data File and Its Description}

The Solr (lucence.apache.org/solr) search server was used.  

\begin{enumerate}
\item The data to be indexed, and supported for search, is put into xml 
with the following structure:

\begin{itemize}
\item \verb+<add>+ and \verb+</add>+ at beginning and end.
\item Each entry defined by \verb+<doc>+ and \verb+</doc>+.
\item A required field providing a unique identifier: 

\verb+<field name=''id''>mm000001102.txt</field>+.
\item Other fields, such as the following for bounding box search:

\verb+<field name=''xcoord''>-0.7341409</field>+ 

\verb+<field name=''ycoord''>-0.09961348</field>+ 

(Note that these coordinates are the principal factor coordinates resulting
from a Correspondence Analysis.  I.e., they are the factor 1 and factor 2
projections, respectively.)

\item Other fields such as: 

\verb+<field name=''name''>''21'' Club Rice Pudding</field>+

\item The main text field, tagged by: \verb+<field name=''recipe''>...</field>+.
\end{itemize}

The xml file was of size over 205.5 MB.  It was put in directory 
\verb+example/exampledocs+.  

\item 
Two files in directory \verb+example/solr/collection1/conf+
need to be fully cognizant of the xml data to be used.  These are
\verb+schema.xml+ and \verb+solrconfig.xml+.  In the latter, changes
made included the following, with regard to \verb+mlt+ or \verb+MLT+,
``More Like This'' option.

In the \verb+requestHandler+ there is: 

\begin{verbatim}
       <arr name="components">
         <str>query</str>
         <str>facet</str>
         <str>mlt</str>
         <str>highlight</str>
         <str>stats</str>
         <str>debug</str>
       </arr>
    </requestHandler>

...

       <!-- Query settings -->
       <str name="defType">edismax</str>
       <str name="qf">
          recipe^1.0
       </str>
       <str name="df">text</str>
       <str name="mm">100%</str>
       <str name="q.alt">*:*</str>
       <str name="rows">10</str>
       <str name="fl">*,score</str>

       <str name="mlt.qf">
          recipe^1.0
       </str>
       <str name="mlt.fl">recipe</str>
       <!-- Following: no. of nearest neigbours returned in MLT -->
       <int name="mlt.count">1</int>

       <!-- Faceting defaults FM CHANGED TO off -->
       <str name="facet">off</str>
\end{verbatim}

\item In file \verb+schema.xml+ the fields used and their definitions 
need to be defined.  This included:

\verb+<field name="recipe" type="text_general" indexed="true" stored="true"+

\verb+termVectors="true"/>+

Also: 
\begin{verbatim}
   <field name="xcoord" type="double" indexed="true" stored="true" />
   <field name="ycoord" type="double" indexed="true" stored="true" />
\end{verbatim}

\item For the browser-based querying, this is management by the 
\verb+velocity+ component, that is in subdirectory 
\verb+example/solr/collection1/conf/velocity+.

The files changed there, to be appropriate for the data analyzed, were: 
\verb+browse.vm, footer.vm, header.vm, product-doc.vm, query.vm+.    Display 
images (see the principal factor plane in Figures \ref{screen1}, \ref{screen2}) 
used there are located in the
named image directory, which is currently \\
\verb+example/solr-webapp/webapp/img+.  

\end{enumerate}

\subsection{Running the Server and Updating the Index}

The Solr server is started thus, in directory \verb+example+, where in this example, 1 GB 
of memory is provided:

\verb+java -Xmx1024m -jar start.jar+

To update, or commence, indexing, in the directory containing the xml data, 
\verb+example/exampledocs+,
the following command is issued.  This supposes a running server.

\verb+java -jar post.jar recipes-F1F2-1.xml+ 

The unique identifier field is the crucial aspect of what gets taken into the index,
or updated.   

\subsection{Querying and Other Operations}

\subsubsection{Web Browser User Interface}

The following access address is used, based on the running server:

\verb+http://localhost:8983/solr/browse+

Through the upper right hand corner link to the Admin screen, or directly using 
\verb+http://localhost:8983/solr/#/collection1+, there is availability of the 
log file; querying can be carried out; statistics of use and of the data can be accessed.
Note that as currently configured in this work, 
\verb+collection1+ contains the indexed data.

Example of query: ``sugar beer pasta''.  A required term is specified in the query with a 
preceding plus sign, and a requirement not to have a term is specified with a minus sign 
preceding the term.  The  ``More Like This'' option 
gives a number of nearest neighbours of the document, and its parameters are defined 
in the settings in \verb+schema.xml+.   

\begin{figure}
\begin{center}
\includegraphics[width=16cm]{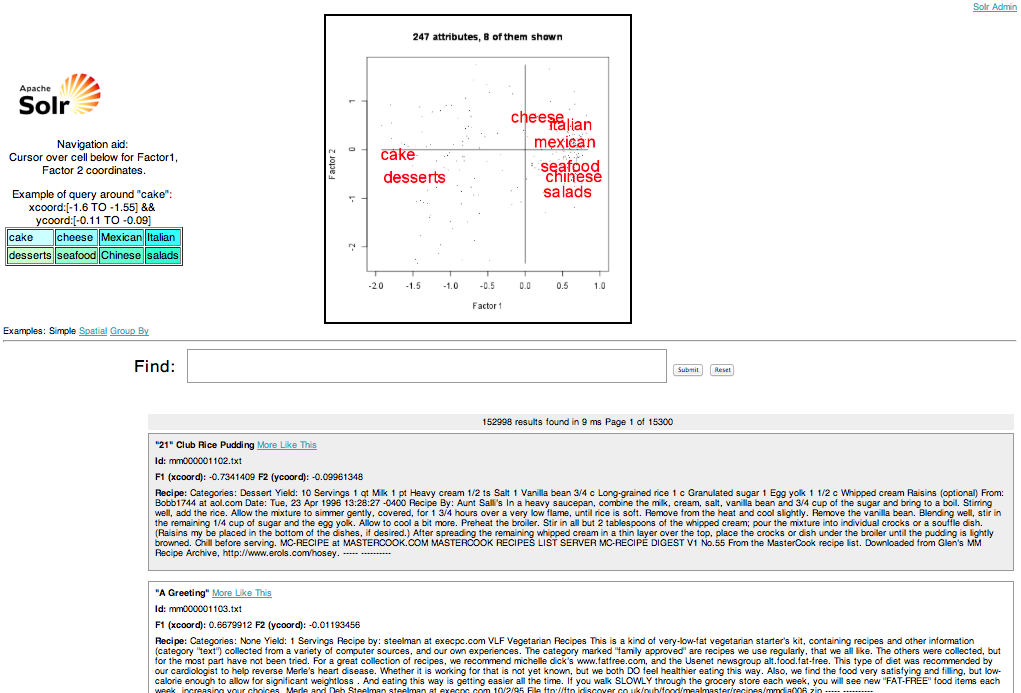}
\end{center}
\caption{Screenshot of browse Solr window, before submitting a query.  For each recipe
(and here, to begin with, the 152,998 recipes are listed), the id (identifier), 
coordinates
(factors 1 and 2), are given, followed by the entire recipe which has had new 
line characters
ignored.   The ``More Like This'' option follows the recipe name.}
\label{screen1}
\end{figure}

\begin{figure}
\begin{center}
\includegraphics[width=16cm]{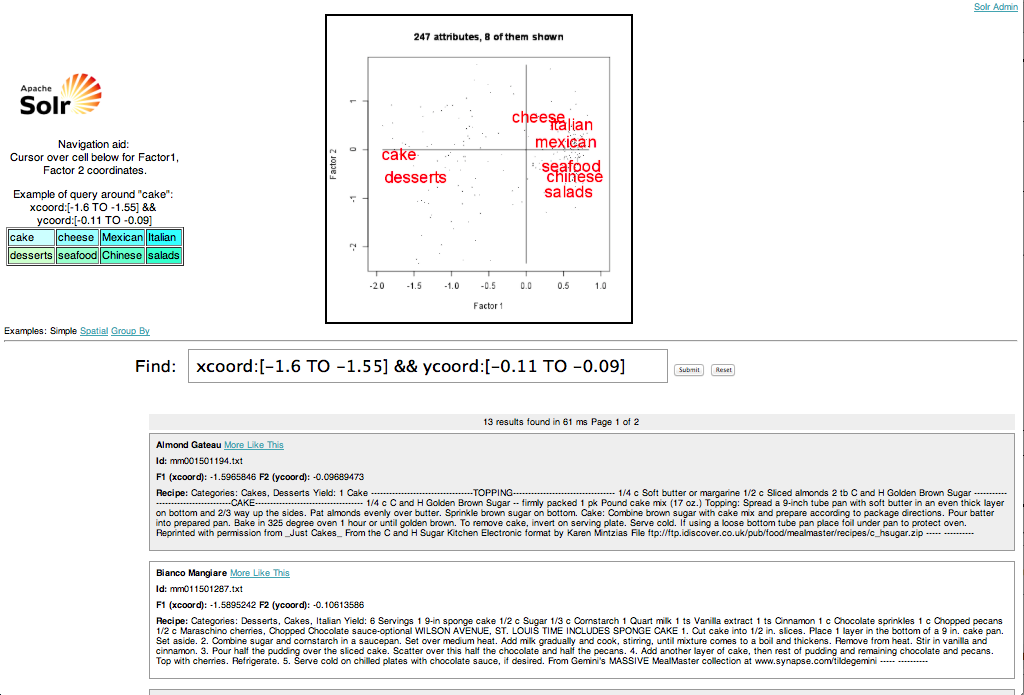}
\end{center}
\caption{Screenshot (second) of browse Solr window. A bounding box query has been 
given, based on the coordinates of the attribute ``cake''.}
\label{screen2}
\end{figure}

Figures \ref{screen1} and \ref{screen2} show examples of use.  The Correspondence 
Analysis figure at the top is static.  It and the table to its left are provided as 
navigation 
aids in search and discovery.  The table to the left of the Correspondence Analysis 
principal 
factor plane is also static.  An alternative could be a text cloud or cloudmap 
(see \cite{cloudmap}). 

\subsubsection{Command Line Querying}

An example follows (all to be placed on one line).  For this, three ``More Like This'' 
near neighbours were required (i.e., in file \verb+solrconfig.xml+, 
there was this setting: \verb+<int name=''mlt.count''>3</int>+).  

\begin{verbatim}
curl -o out.xml 'http://localhost:8983/solr/collection1/
      browse?&q=id:mm078001428.txt&wt=xml&mlt=true' 
\end{verbatim}

In this case, results are saved to file \verb+out.xml+.  From a search 
for recipe \verb+mm078001428.txt+, a ``More Like This'' request is 
submitted.  

From the file \verb+out.xml+ here is a little utility to write out 
just the recipe identifiers (all on one line):

\begin{verbatim}
awk -v srch="\"id\">" 'BEGIN{l=length(srch)} END{print "\n"} 
{t=match($0,srch);if(!t){next}extr=substr($0,t+5,15);printf "%s ", extr}' 
< out.xml
\end{verbatim}

In this case, this gives:

\begin{verbatim}
mm078001428.txt mm110501451.txt mm158501305.txt mm161501159.txt 
\end{verbatim}

\section{Consistency of Solr Searches}

Using identifier (and recipe) ``mm078001428.txt'', with name 
``No Fat, No Salt, No Sugar Vanilla Ice Cream'',   
we found its ``More Like This'' 
best match to be ``mm110501451.txt'', with name 
``Sugar-Free Cappuccino Ice Cream''.  Using the latter, 
we found its best match to be the former, thereby showing consistency.
More specifically, the two recipes in this case were mutual 
or reflexive nearest neighbours.  See \cite{mur85a} for the use of 
this principle (of mutual nearest neighbours, and also nearest 
neighbour chains) in agglomerative hierarchical clustering algorithms.  

We also sought best match recipes using the Correspondence Analysis 
factor space.  This latter is of course Euclidean.  We used the full 
space dimensionality (hence 247 less 1 due to linear dependence through 
centring the cloud of recipes, and the dual cloud of ingredients).   
For recipe ``mm078001428.txt'', we got its closest neighbour as
``mm048501554.txt'', ``Fruit Flavor Milk Shakes''.   For recipe 
``mm110501451.txt'', we got its closest neighbour as 
``mm161501159.txt'', ``Vanilla Ice Cream - Diabetic *WW No.2'' 

\begin{figure}
\begin{center}
\includegraphics[width=10cm]{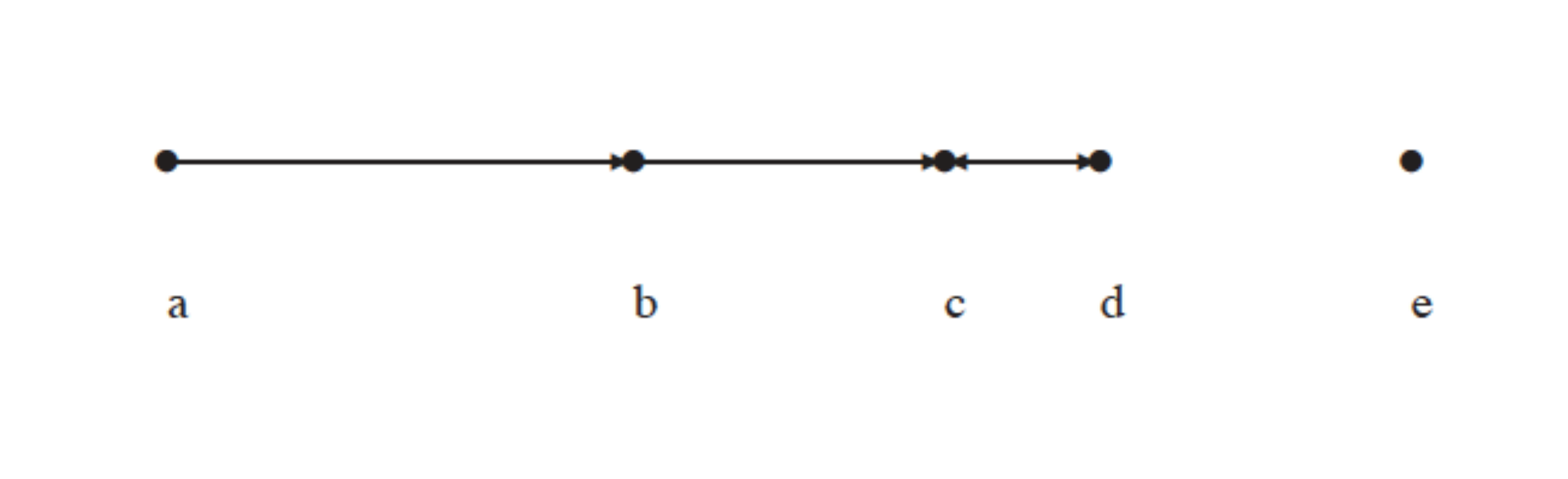}
\end{center}
\caption{Nearest neighbours.  Point $b$ is the nearest neighbour of point
$a$.  
Point $c$ is the nearest neighbour of point $b$.  Point $d$ is the nearest
neighbour of point $c$, and reciprocally $c$ is the nearest neighbour of point
$d$.}
\label{NNs}
\end{figure}

In Figure \ref{NNs} the geometric situation is depicted, 
irrespective of how nearest neighbour is defined
(it need not be a distance but some dissimilarity satisfying $d(a,b) = 
0 \mbox{ if } a = b$; $d(a, b) \geq 0$; and 
the closer $a$ is to $b$, i.e.\ the
more alike they are, the smaller the value of $d(a,b)$).   

The squared Euclidean distance in the full dimensional Euclidean
factor space gives for the pair of best match recipes furnished 
by Solr, 18.94413.  Then in the full dimensional factor space, 
the best match distance squared of ``mm078001428.txt'', as noted 
above, was 4.934411; and the best match distance squared of 
``mm110501451.txt'' was 6.103886.  To note that the Correspondence
Analysis best match distances are supported by the data is to be 
unfair to what is, with Solr, a different framework for determining 
best matches.  This includes using all words, with stemming and 
lemmatization, compared to the 247 ingredients used in the Correspondence
Analysis case.  

We note that the Solr best match information lends itself to 
looking for clusters such that nearest neighbour chains are followed,
and once a (mutual or reciprocal) nearest neighbour pair is found,
they can be agglomerated with no impact on close-by recipes.  This is 
due to Bruynooghe's reducibility property.  Background on nearest
neighbour chain, and reciprocal nearest neighbour based hierarchical
agglomerative clustering can be found in \cite{mur85a}.

\section*{Appendix 1: Sample Recipe}

Recipe mm160001 collection, number 245 out of 500 in sequence from this collection, 
is as follows. 

\begin{verbatim}
 
      Title: The Perfect Roast Chicken with Roasted Shallots And Portob
 Categories: Lifetime tv, Life5
      Yield: 4 servings
 
      4 lb Young roaster
           Salt and pepper; to taste
      1 md Onion; halved and peeled
      3    Cloves garlic; peeled and
           -smashed
      1 bn Fresh herbs; rosemary thyme
           ; flat-leaf parsley
    1/4 c  Olive oil
      2 c  Chicken broth; divided
      8 lg Shallots
      8    Portobello mushrooms; stems
           -removed and
           ; cut in half
    1/2 c  Dry white wine
 
  How to Prepare the Chicken:
  
  1. Preheat the oven to 350 F. Season the skin and the inside of the cavity
  of the chicken generously with salt and pepper.
  
  2. Place the onion, garlic, and fresh herbs in the cavity and truss the
  chicken. In a roasting pan over medium-high heat, heat the oil until it
  begins to smoke.
  
  3. Place the chicken, breast-side-down, in the oil and cook until all the
  sides off the chicken are completely browned.
  
  4. Place the bird, breast-side-up into the oven and baste with 1/4 cup of
  the chicken stock. Continue basting with 3/4 cup of the chicken stock every
  10 minutes until the chicken is done, approximately 1 hour, or until the
  chicken reaches an internal temperature of 170 F.
  
  5. Add the shallots 20 minutes after the chicken has been put into the oven
  and the mushrooms 40 minutes after the chicken has been put into the oven.
  
  6. Remove the cooked chicken, shallots, and mushrooms from the oven and set
  on a platter, cover and let rest for 10 minutes.
  
  How to Prepare the Sauce:
  
  1. Place the roasting pan back on the stove over medium-high heat and bring
  the juices in the pan to a boil. Using a small ladle remove any of the fat
  that has risen to the top of the pan.
  
  2. Add the white wine to the pan and, using a wooden spoon, scrape up the
  browned bits from the bottom of the pan. Reduce the wine by half and add
  the remaining 1 cup of chicken stock.
  
  3. Reduce the stock by half, season with salt and pepper and add the fresh
  thyme. Carve the chicken and arrange on a platter with the shallots and
  mushrooms. Spoon the pan juices over the meat.
  
  The Perfect Roast Chicken With Roasted Shallots and Portobello
  
  <A9> 1997 Lifetime Entertainment Services. All rights reserved.
  
  MC formatted using MC Buster by Barb at PK
  
    Converted by MM_Buster v2.0l.
 
-----
\end{verbatim}

We note the following in this recipe text: 
Abbreviations (``lg'', large; ``lb'', pounds; ``c'', cup; etc.).  
Source and 
other summarized or abbreviated data.   A control character (``$<$A9$>$''). 

\section*{Appendix 2: 247 Ingredients Used as Attributes in the Correspondence 
Analysis}

List of 247 ingredients used, with their frequencies of occurrence. 

{\small 
\begin{verbatim}
          salt          sugar          water         pepper            oil         butter 
        167636         142624         134454         132012         126242         115387 
         sauce          flour         garlic          cream         cheese        chicken 
        105909          99870          85080          83859          80835          78635 
         onion          juice            egg           milk          lemon           eggs 
         73933          65773          62000          61125          55408          50086 
         bread         onions           rice      chocolate          olive        vanilla 
         48789          44169          41645          40737          39539          38386 
          cake       tomatoes        parsley       potatoes        vinegar      vegetable 
         35317          34894          32217          31423          31340          30735 
          wine         tomato          beans           beef     vegetables         cloves 
         29192          28747          28180          28761          27203          27034 
          soup         orange       cinnamon      margarine      mushrooms          salad 
         26558          25976          25656          24266          23693          23612 
          fish           corn          broth         celery        mustard            pie 
         23267          21212          20832          21193          20192          20081 
         pasta          fruit        peppers            soy          chili          basil 
         19743          18989          18695          18036          17148          16354 
        shrimp           soda        cookies          syrup        carrots          honey 
         16332          16257          16216          15652          15197          14733 
        sodium         cookie       parmesan            ice       dressing     cornstarch 
         14034          14173          13878          13928          13886          13848 
         thyme          bacon         pastry           lime          yeast         potato 
         13636          13653          13309          13274          13347          12957 
         apple        protein        spinach      casserole        oregano          cumin 
         12642          12129          11889          12106          11855          11996 
        nutmeg    cholesterol        raisins          clove        coconut      pineapple 
         11756          11376          11019          11042          10892          10864 
         roast          chips           chop          puree        topping       marinade 
         10758          10667          10637          10638          10542          10586 
       noodles           loaf       desserts       cilantro          yolks         peanut 
         10441          10456          10371          10405          10362          10258 
       italian          chile      seasoning         apples        almonds           peas 
         10285          10293          10350          10280          10216          10459 
        sesame         turkey            ham        cabbage        paprika           leaf 
         10184           9975           9725           9736           9689           9703 
         mixer         yogurt      coriander   carbohydrate        sausage        cayenne 
          9211           9067           8903           8765           8918           8892 
          lamb          wheat           bean        walnuts          cakes           mint 
          8571           8639           8573           8470           8398           8256 
       lettuce     mayonnaise         pecans         sherry          cocoa        pudding 
          8159           8000           7893           7950           7835           7843 
       cheddar          grain         salmon         olives         carrot          shell 
          7877           7806           7724           7744           7767           7511 
    vegetarian       broccoli       zucchini          salsa         flakes         grease 
          7405           7413           7290           7268           7104           7079 
       chinese       shallots        poultry       mushroom          steak       rosemary 
          7079           6877           6969           6857           6927           6699 
      eggplant         chiles           rind          curry         coffee           dill 
          6708           6804           6696           6798           6602           6683 
        spices         breads     buttermilk worcestershire         starch        seafood 
          6648           6670           6351           6273           6031           6135 
       pumpkin        gelatin  carbohydrates      scallions         almond         chives 
          5820           5899           5749           5815           5662           5621 
         spice          meats          herbs           tofu        dessert          pizza 
          5576           5591           5496           5409           5466           5398 
  strawberries         juices        muffins        mexican      tortillas          chops 
          5373           5318           5290           5176           5139           4996 
           rum          icing          soups       cornmeal         emeril      asparagus 
          5014           4917           5084           5032           4874           4740 
        sauces         muffin         fruits       stuffing          jelly         salads 
          4792           4778           4737           4704           4756           4731 
        shells       jalapeno     mozzarella         banana         steaks         greens 
          4645           4581           4484           4387           4354           4390 
     spaghetti        broiler       cucumber       cherries          toast         cherry 
          4343           4346           4679           4351           4307           4358 
       chilies           yolk          gravy          cider           root        tabasco 
          4324           4289           4258           4244           4304           4272 
          oats          fiber           veal       tortilla           sage          dijon 
          4225           4135           4061           4085           4040           4040 
       bananas          broil           tuna       molasses        peaches    peppercorns 
          3970           3970           3914           3949           3939           3928 
         candy           duck       tarragon         fennel           tart        custard 
          3884           3805           3851           3789           3814           3774 
         maple       scallops           stew         brandy        berries          pears 
          3749           3871           3802           3780           3777           3559 
          crab  confectioners       crackers       biscuits       bouillon        peanuts 
          3611           3571           3570           3464           3526           3440 
         leeks           beer       turmeric         stalks        ricotta        oranges 
          3392           3450           3385           3393           3274           3278 
       lentils      raspberry        cracker           herb    raspberries     strawberry 
          3260           3216           3128           3113           3083           3093 
           jam 
          3151 
\end{verbatim}
}

\section*{Appendix 3: Alternative Presentation of Plot}

\begin{figure}
\begin{center}
\includegraphics[width=16cm]{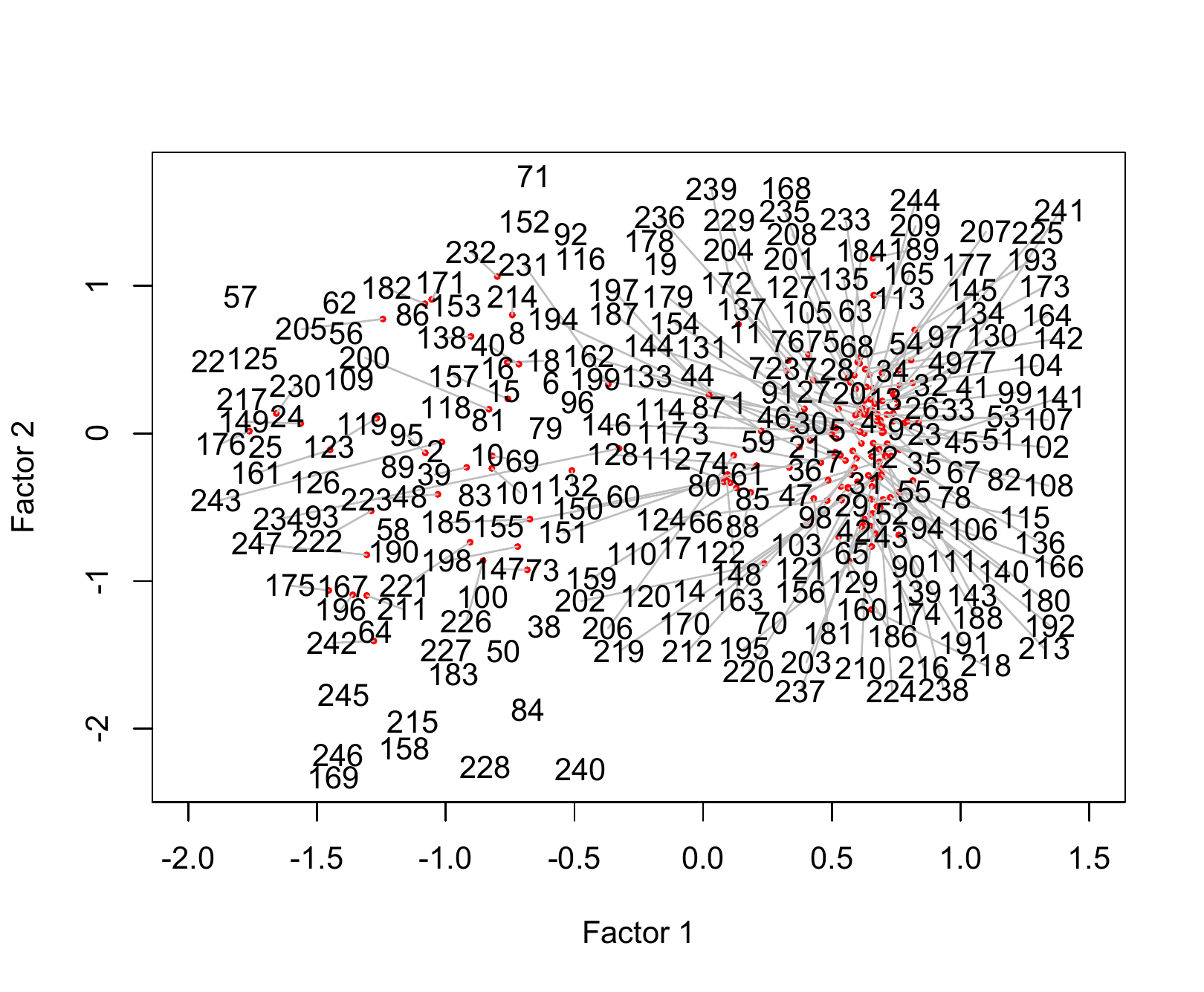}
\end{center}
\caption{Principal factor plane showing projections of 152,998 recipes,
each represented by a dot.}
\label{textplot}
\end{figure}

Figure \ref{textplot} shows an improved presentation of 
Figure \ref{F1F2ing}, through movement 
of projections (from the red dots, indicated by the stick lines), as
implemented in the {\tt textplot} program in the Word Cloud package 
in R \cite{fellows}.  

\section{Appendix 4: Correspondence Analysis for Singular/Plural Association and 
Potentially for Disambiguation}

In the recipe set, there were 101,060 unique words.  Words were as defined 
by us: length $\geq 1$; all upper case set to lower case; punctuation 
removed; numeric and accented characters ignored.  Variations on spelling
included the following example, with numbers of occurrence in the recipe 
set:
\begin{verbatim} 
zucchini 7155
zuchinni 86
zuccini 57
zucchinis 56
zuchini 40
zucchine 14
zucchinni 7
zucchin 6
zuckinni 5
zuchinnis 4
zuccinni 3
zuccchini 2
zuccihini 2
zuchhini 2
azucchini 1
brushingzucchini 1
lambzucchini 1
ofzucchini 1
pzucchini 1
zucchiniabout 1
zucchinii 1
zucchinnis 1
zucchni 1
zucchnini 1
zuccinis 1
zuchhinis 1
zuchine 1
zuchinis 1
zucinni 1
zzzucchini 1
\end{verbatim}

On this list we see some words run together, singulars and plurals, but 
also all manner of misspellings.   

We examine here, using the 152,998 $\times$ 247 ingredients set, the 
projections in the principal factor plane of singulars and plurals. 
Recall that the principal factor plane, while being the best visualization 
of the data, with inherent dimensionality 246, only accounts for just over
3.8\% of the total inertia.  

The singulars used were as follows, with additionally their 23 
corresponding plurals:
``strawberry'', ``carrot'', ``muffin'', ``egg'', ``apple'', ``yolk'',
``tortilla'', ``bread'', ``cookie'',
   ``pepper'', ``juice'', ``vegetable'', ``clove'', ``carbohydrate'',
 ``spice'', ``soup'', ``peanut'', ``onion'', ``olive'',
   ``cracker'', ``almond'', ``bean'', ``banana''.

\begin{figure}
\begin{center}
\includegraphics[width=16cm]{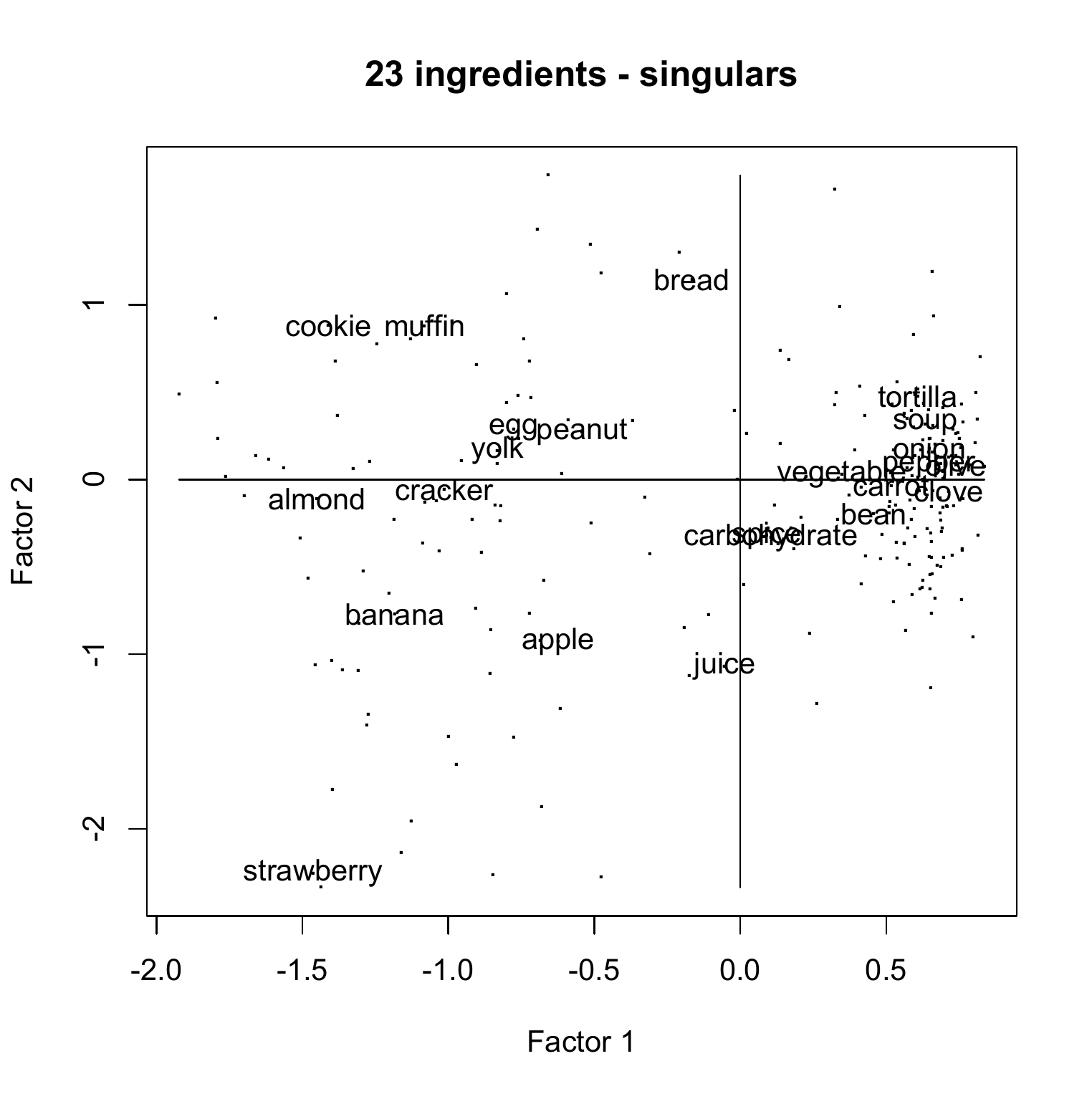}
\end{center}
\caption{Principal factor plane showing projections of 247 ingredients,
using a dot to represent projection, together with 23 singulars of 
ingredient names.  Cf.\ Figure \ref{plurs} for the corresponding
plurals of these names.}
\label{sings}
\end{figure}

\begin{figure}
\begin{center}
\includegraphics[width=16cm]{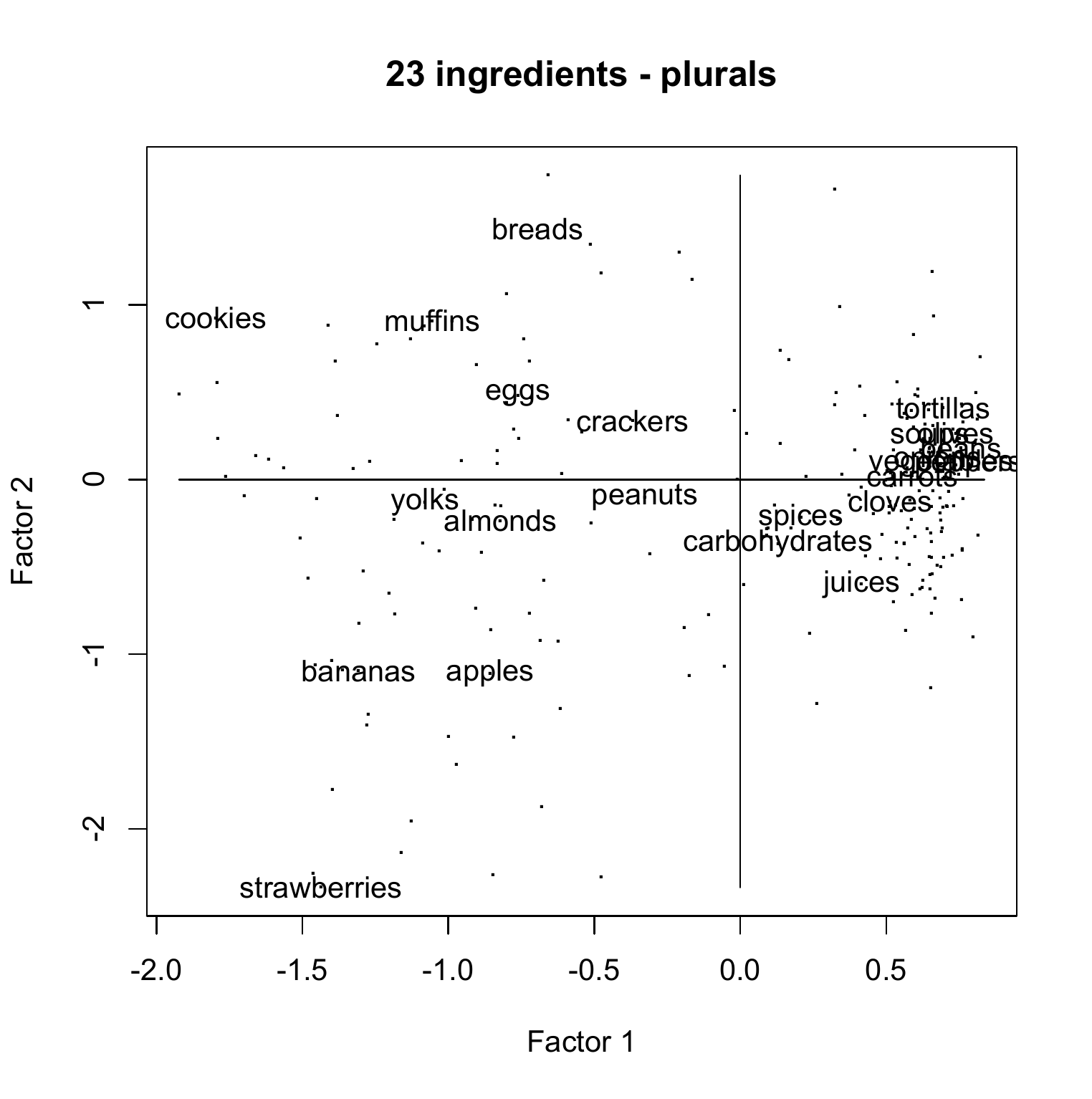}
\end{center}
\caption{Principal factor plane showing projections of 247 ingredients,
using a dot to represent projection, together with 23 plurals of 
ingredient names.  Cf.\ Figure \ref{sings}.}
\label{plurs}
\end{figure}

\begin{figure}
\begin{center}
\includegraphics[width=16cm]{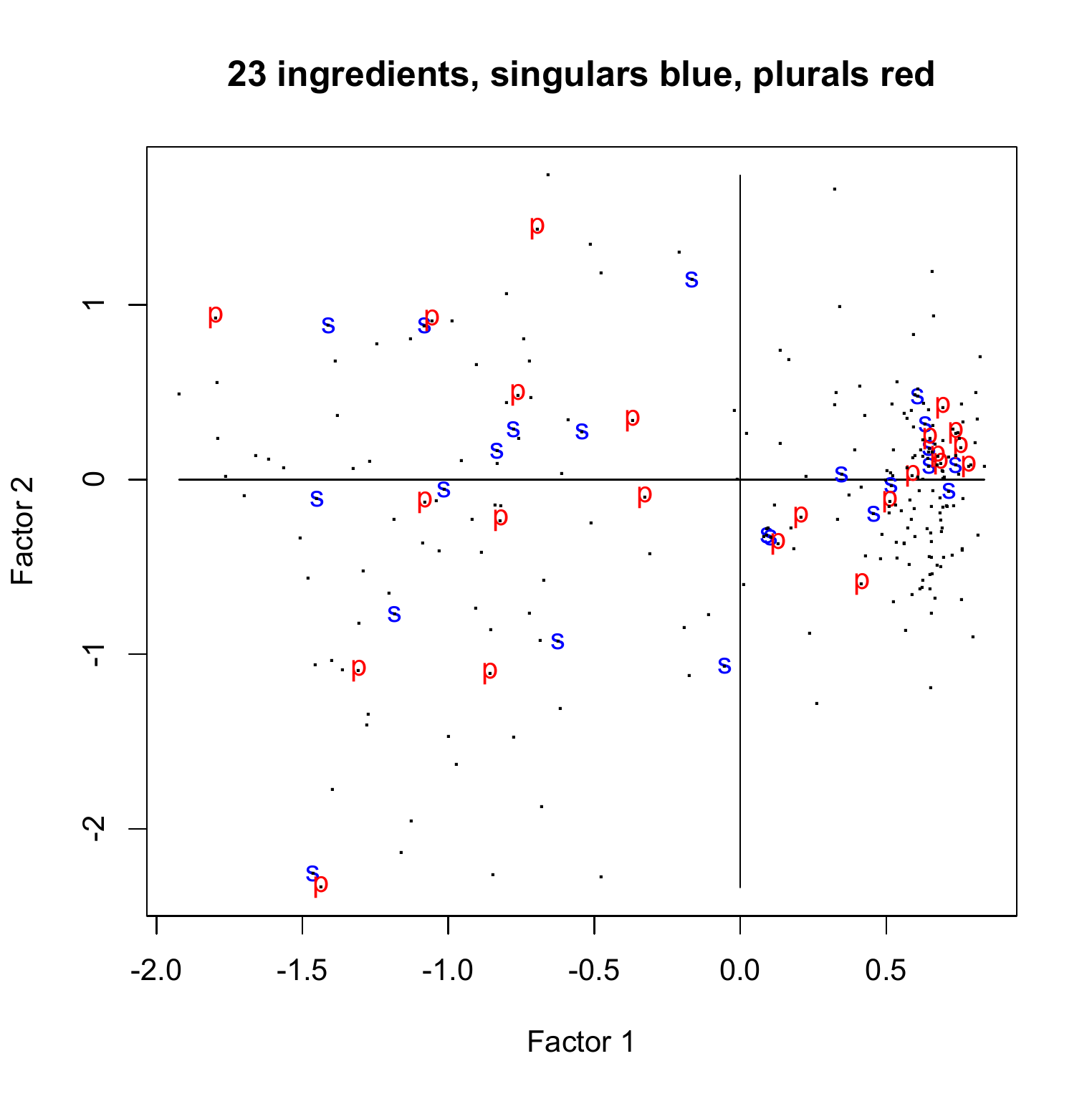}
\end{center}
\caption{Combining Figures \ref{sings} and \ref{plurs}, 
this shows 23 singulars (``s'') and plurals (``p'').}
\label{sandp}
\end{figure}

\begin{figure}
\begin{center}
\includegraphics[width=16cm]{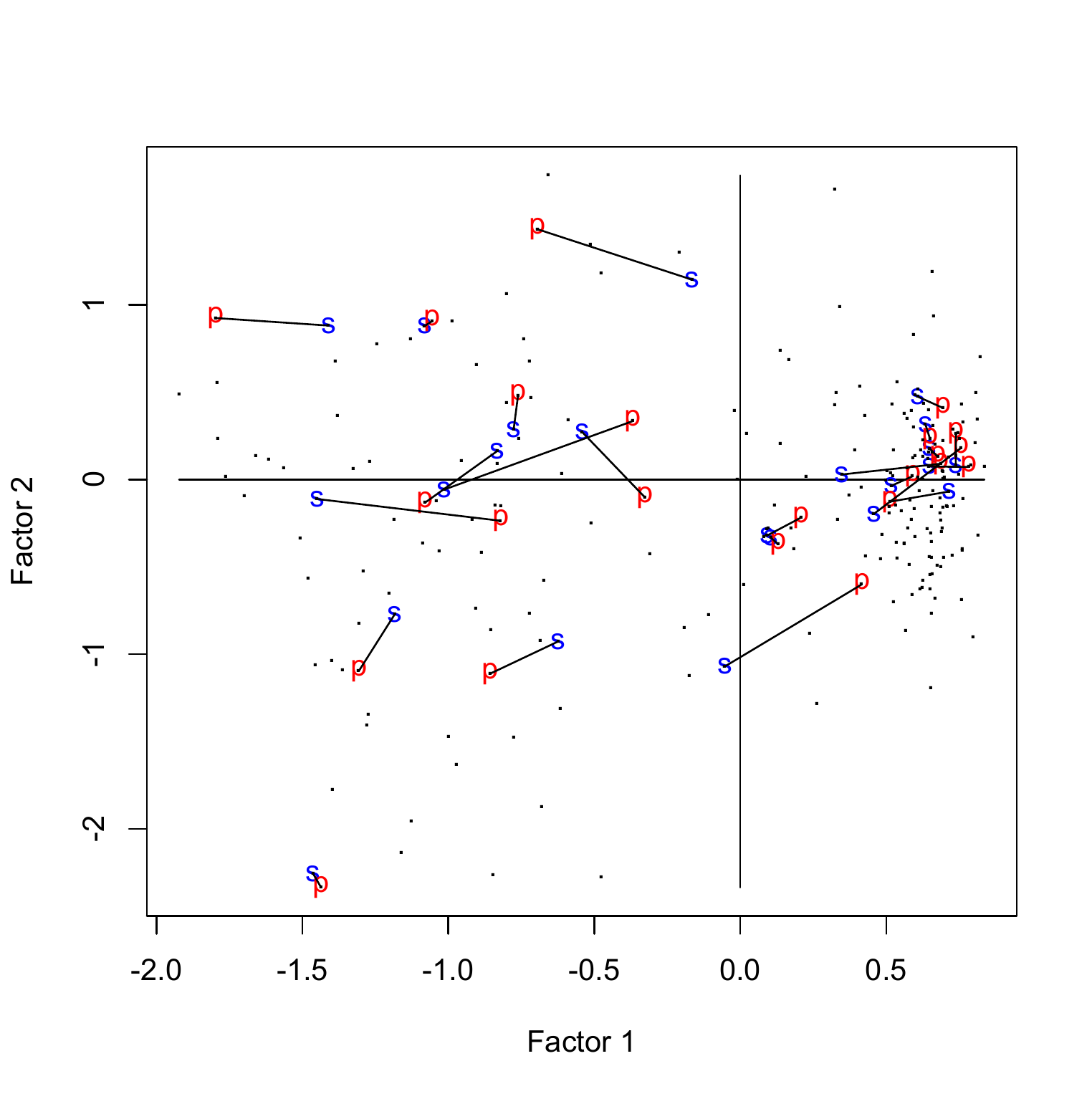}
\end{center}
\caption{Links shown for all singular and plural pairs.}
\label{sandplinks}
\end{figure}

Figures \ref{sings}, \ref{plurs}, and also \ref{sandp} with links
drawn for all singular and plural pairs, show the
close association in the principal factor plane.  Some singulars
and plurals are admittedly somewhat separated, e.g.\ ``cracker'',
``yolk'', pointing to some difference in semantic -- clearly 
possible in practice -- and contextual 
use of singular and plural.  Overall the association between the 
terms is well exemplified in the figures.  

\end{document}